\documentclass[conference]{IEEEtran}
\IEEEoverridecommandlockouts
\usepackage{cite}
\usepackage{amsmath,amssymb,amsfonts}
\usepackage{algorithmic}
\usepackage{graphicx}
\usepackage{textcomp}
\usepackage{xcolor}
\usepackage{siunitx}
\usepackage{booktabs}
\usepackage{subcaption}
\usepackage{tabularx}
\usepackage{url}
\usepackage[absolute,overlay]{textpos}
\usepackage{caption}
\usepackage[absolute,overlay]{textpos}
\usepackage{etoolbox}
\makeatletter
\patchcmd{\@maketitle}{\addvspace{0.5\baselineskip}\egroup}{\addvspace{-1\baselineskip}\egroup}{}{}
\makeatother

\def\BibTeX{{\rm B\kern-.05em{\sc i\kern-.025em b}\kern-.08em
    T\kern-.1667em\lower.7ex\hbox{E}\kern-.125emX}}
\begin{document}

\title{K-AoI: A Spatio-Temporal Synchronization Metric for Digital Twin-based Teleoperation in Dynamic Network Environments\\

\thanks{For the purpose of open access, the author has applied a Creative Commons Attribution (CC BY) license to any Author Accepted Manuscript version arising.}
}

\author{\IEEEauthorblockN{Lu Tian}
\IEEEauthorblockA{\textit{School of Engineering} \\
\textit{University of Warwick}\\
Coventry, UK \\
lu.tian.2@warwick.ac.uk}
\and
\IEEEauthorblockN{Zhenhui Yuan}
\IEEEauthorblockA{\textit{School of Engineering} \\
\textit{University of Warwick}\\
Coventry, UK \\
zhenhui.yuan@warwick.ac.uk}
\and
\IEEEauthorblockN{Zexin Deng}
\IEEEauthorblockA{\textit{School of Engineering} \\
\textit{University of Warwick}\\
Coventry, UK \\
zexin.deng@warwick.ac.uk}
\and\textbf{}
\IEEEauthorblockN{Wenjuan Zhang}
\IEEEauthorblockA{\textit{Warwick Business School} \\
\textit{University of Warwick}\\
Coventry, UK \\
wenjuan.zhang@wbs.ac.uk}
}

\maketitle

\begin{textblock*}{\textwidth}(0.62cm,25.9cm)

\end{textblock*}

\begin{abstract}
Synchronization between a digital twin (DT) and its physical robot is critical for effective teleoperation, as mismatches directly degrade operator awareness and task performance. Existing synchronization metrics are predominantly time-based and rarely capture the spatial consequences of delayed information during motion. In this paper, we introduce \emph{Kinematic Age of Information} (K-AoI), a spatio-temporal metric that weights AoI by instantaneous Cartesian velocity, translating temporal staleness into physically meaningful spatial divergence. We implement and evaluate K-AoI on a DT teleoperation testbed comprising a six-degrees-of-freedom (6-DOF) manipulator, a Unity-based twin, and controlled network impairments using \texttt{tc netem} across seven network conditions. Per-sample analysis shows that K-AoI exhibits rank-correlations with AoI at $\rho \le 0.65$ across all networks, dropping to $\rho = 0.22$ under satellite connections. This shows that K-AoI is not a simple transformation of AoI, but instead captures velocity-dependent degradation effects that time-only metrics cannot resolve. We further analyze AoI tail behavior using Ultra-Low AoI (ULAoI) and find that the commonly assumed Generalized Pareto Distribution (GPD) model is rejected. This reflects the impact of deterministic forwarding and bounded jitter, which violate standard extreme-value assumptions. Overall, K-AoI provides a complementary and physically grounded synchronization metric, linking network delay to task-relevant spatial error in digital-twin teleoperation systems. The dataset and testbed software are publicly available at: \texttt{\url{https://github.com/luziyaotian/kaoi-testbed}}.
\end{abstract}


\vspace{-1.6\baselineskip}

\section{Introduction}
Robot teleoperation allows human control of physical systems in hazardous or inaccessible environments~\cite{hri2020latency}. A key challenge lies in the network connection between the human operator and the robot, as delays exceeding approximately \SI{250}{ms} can significantly degrade situational awareness and task performance~\cite{hri2020latency}. Digital twins (DT) mitigate this by maintaining a synchronized virtual representation, allowing the operator to interact with a local, low-latency virtual robot while control commands are transmitted to the physical robot~\cite{robotwin2025, constructwin2026}.
However, evaluating the performance of DT-based teleoperation systems remains an open challenge. Subjective workload measures such as NASA-TLX~\cite{robotwin2025} cannot be derived from system logs and do not generalize well across different operators. Latency reports typically reflect average behavior and miss tail spikes that cause task failure. Age of Information (AoI)~\cite{kaul2012aoi} addresses temporal freshness more explicitly, while Ultra-Low AoI (ULAoI)~\cite{liao2023ulaoi} extends this idea by modeling tail behavior using extreme-value theory.
However, both AoI and ULAoI are purely time-based metrics and ignore physical motion: a \SI{500}{ms} delay has negligible impact on a stationary DT, yet produces \SI{250}{mm} of spatial divergence between the robot end-effector and its digital twin at \SI{500}{mm/s}. Since manipulation performance is fundamentally spatial, and the critical error is the Cartesian distance between the physical robot and its digital counterpart, existing metrics do not capture this combined effect of temporal delay and physical motion.
AoI and its extensions quantify information freshness in the temporal domain. The closest related spatial formulation is Age of Positioning (AoP)~\cite{hussain2026aop}, a mean squared error-based (MSE) metric derived analytically for mobile agents under Random Waypoint mobility and queuing-discipline updates. Our work shares the idea of incorporating motion into a freshness-aware metric, but differs in setting (fixed-based teleoperation versus stochastic mobility), methodology (empirical testbed versus closed-form), and form (linear scale distance versus quadratic MSE). Spatial AoI~\cite{spatialAoI2023} also extends AoI spatially, but focuses on vehicular networks rather than robotic manipulation. On the DT side, RoboTwin~\cite{robotwin2025}, Huang \textit{et al.} \cite{huang2025multisensory} and ConstrucTwin~\cite{constructwin2026} are system-level contributions without a motion-aware metric. Kawshan \emph{et~al.}~\cite{kawshan2026unityros} and the network-aware TeleSim benchmark~\cite{deng2025telesim} report latency mean and task-performance statistics on comparable teleoperation testbeds, but neither analyze the tails nor spatial divergence. Chen \emph{et~al.}~\cite{chen2026gsc} treats reconstruction error as a constraint while compressing transmission (35--44\% load reduction), whereas we treat it as the quantity of interest.
Roy \emph{et~al.}~\cite{roy2025qoc} address resource dimensioning analytically through quality of control bounds rather than empirical measurement on a specific testbed.
This paper makes the following contributions:
\begin{enumerate}
\item \textbf{A spatio-temporal synchronization metric for digital-twin teleoperation.}
      We introduce Kinematic Age of Information (K-AoI), a velocity weighted extension of AoI that converts temporal staleness into a physically meaningful spatial divergence, explicitly capturing when network delay translates into task error during motion.
\item \textbf{A reproducible experimental framework for digital twin teleoperation under network degradation}, integrating physical robotics, real-time simulation, and controlled network impairment, to enable systematic evaluation of synchronization and communication metrics across different network conditions.
\item \textbf{Empirical validation over ($N=11$ trials)} showing that K-AoI captures
      velocity degradation effects not shown in AoI. Across all network profiles, the Spearman rank correlations per-sample between AoI and K-AoI remain below $\rho = 0.65$, decreasing to $\rho = 0.22$ under satellite conditions, indicating clear divergence under high-latency regimes.
\end{enumerate}

\begin{table}[t]
\caption{K-AoI positioned against prior synchronization metrics
for networked robotic systems.}
\label{tab:related}
\centering
\footnotesize
\setlength{\tabcolsep}{4pt}
\renewcommand{\arraystretch}{1.0}
\begin{tabular}{@{}llccc@{}}
\hline
\textbf{Work} & \textbf{Metric} & \textbf{Spatial} & \textbf{Tails} & \textbf{Method} \\ \hline
AoI~\cite{kaul2012aoi}              & Time                & --        & --        & Theory \\
ULAoI~\cite{liao2023ulaoi}          & Time + GPD          & --        & \checkmark & Sim. \\
AoP~\cite{hussain2026aop}           & Position MSE        & \checkmark & --       & Theory \\
Kawshan~\cite{kawshan2026unityros}  & Latency             & --        & --        & Emp. \\
Chen~\cite{chen2026gsc}             & Angle + velocity err.    & \checkmark   & --   & Sim.+Emp. \\
\textbf{K-AoI (ours)}               & \textbf{Time$\times$vel.} & \checkmark & \checkmark & \textbf{Emp.} \\ \hline
\multicolumn{5}{@{}p{0.92\columnwidth}@{}}{\scriptsize \textbf{Spatial:} metric incorporates position/motion. \textbf{Tails:} explicit distributional tail analysis. \textbf{Method:} Theory / Sim(ulation) / Emp(irical testbed).}
\end{tabular}
\end{table}

\vspace{-0.9\baselineskip}
\section{Problem Formulation}
\label{sec:metrics}

We discuss three metrics for DT synchronization, progressing from the standard temporal definition used in IoT through to a velocity-aware extension. Let $g(t)$ denote the generation timestamp of the most recent update received at time $t$.

\subsection{Standard AoI}
Age of Information quantifies information freshness as the elapsed time since the receiver last obtained a new state update~\cite{kaul2012aoi}:
\begin{equation}
    \Delta(t) \;=\; t - g(t)
\end{equation}
While AoI is widely used for evaluating networked control systems, it is typically reported using time-averaged means or percentile summaries. However, in DT teleoperation, such aggregate temporal measures are insufficient. Two systems may exhibit identical mean AoI values while having fundamentally different delay distributions, particularly in their tail behavior. It is these rare but extreme latency spikes, rather than average delays, that are often responsible for degradation in operator performance and eventual task failure.

\subsection{Ultra-Low AoI (ULAoI)}
\label{sec:ulaoi}
Following Liao \emph{et~al.}~\cite{liao2023ulaoi}, for a freshness threshold $\tau$ we characterize the AoI tail through:
\begin{align}
P_\tau &\;=\; \Pr\!\left[\Delta > \tau\right]
\label{eq:pviol}\\
E_\tau &\;=\; \mathbb{E}\!\left[\Delta - \tau \;\middle|\;\Delta > \tau\right]
\label{eq:excess}
\end{align}
$P_\tau$ is the probability of crossing the freshness bound; $E_\tau$ is the conditional mean excess above the bound. We set $\tau = \SI{200}{ms}$, which is consistent with thresholds at which teleoperation studies report measurable degradation in operator performance and task completion time~\cite{kim2005telerobotic}. We calculate $P_\tau$ and $E_\tau$ empirically, and we additionally evaluate the GPD parametric form proposed by Liao \emph{et~al.}~\cite{liao2023ulaoi} in Section~\ref{sec:gpd-fit}.

\subsection{Why time alone is insufficient}
Both AoI and ULAoI are time-based metrics, yet in robotic teleoperation, the impact of stale information depends not only on delay duration but also on the system’s motion during that period. If both the operator and robot are stationary, even a \SI{1}{s} delay causes no spatial divergences. In contrast, when the system is moving at \SI{500}{mm/s}, a \SI{200}{ms} delay causes a \SI{100}{mm} gap between the DT and real end-effector positions. This motivates a metric integrating information freshness with physical motion, quantifying when temporal delays translate into meaningful spatial errors.

\section{K-AoI: Kinematic Age of Information}
\label{sec:kaoi}
Building on the idea of weighting AoI by motion~\cite{hussain2026aop, spatialAoI2023}, we define Kinematic Age of Information (K-AoI) for a tracked entity with instantaneous Cartesian velocity $\mathbf{v}(t) \in \mathbb{R}^{3}$:
\begin{equation}
\mathrm{K\text{-}AoI}(t) \;=\;
\Delta(t) \cdot \left\| \mathbf{v}(t) \right\|_{2}
\label{eq:kaoi}
\end{equation}
K-AoI is expressed in units of length, thereby translating temporal information staleness into a physically meaningful spatial metric. Under the assumption of locally constant velocity over the staleness interval $\Delta$(t) window, K-AoI provides a first-order approximation of the Cartesian divergence between the physical end-effector and its DT. In scenarios involving non-linear motion, the approximation error is bounded and typically reduced due to the short update intervals in continuous teleoperation systems, making K-AoI suitable for real-time evaluation.
For end-effector analysis, the velocity is calculated by applying the forward kinematics to the measured joint states and finite-differencing the resulting Cartesian trajectory. We denote this variant as $\mathrm{K\text{-}AoI}_\mathrm{eef}$. We use this Cartesian formulation rather than using the joint angle or velocity errors used in compression-oriented DT work~\cite{chen2026gsc}, since the spatial gap at the end-effector, not in the joint configuration space, is what operators perceive in manipulation tasks.
Unlike AoP~\cite{hussain2026aop}, which is derived analytically under stochastic mobility and queuing assumptions, K-AoI is directly computed from streamed robot state measurements without probabilistic mobility models or queuing-theoretic constructs, making it applicable to DT teleoperation systems independent of network or traffic assumptions.

\section{Testbed and Protocol}
\label{sec:testbed}

\subsection{Testbed Implementation}
\begin{figure}[t]
\centering
\includegraphics[width=\columnwidth]{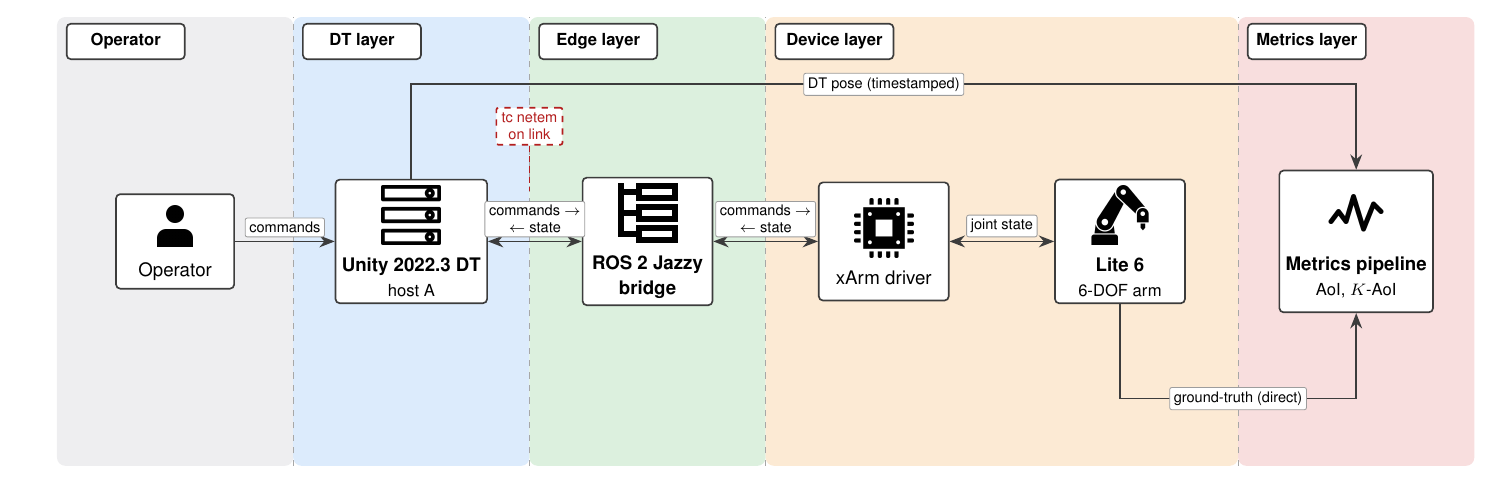}
\caption{Testbed architecture}
\label{fig:testbed}
\end{figure}

\begin{figure}[t]
\centering
\begin{subfigure}[b]{0.42\columnwidth}
    \centering
    \includegraphics[width=\linewidth]{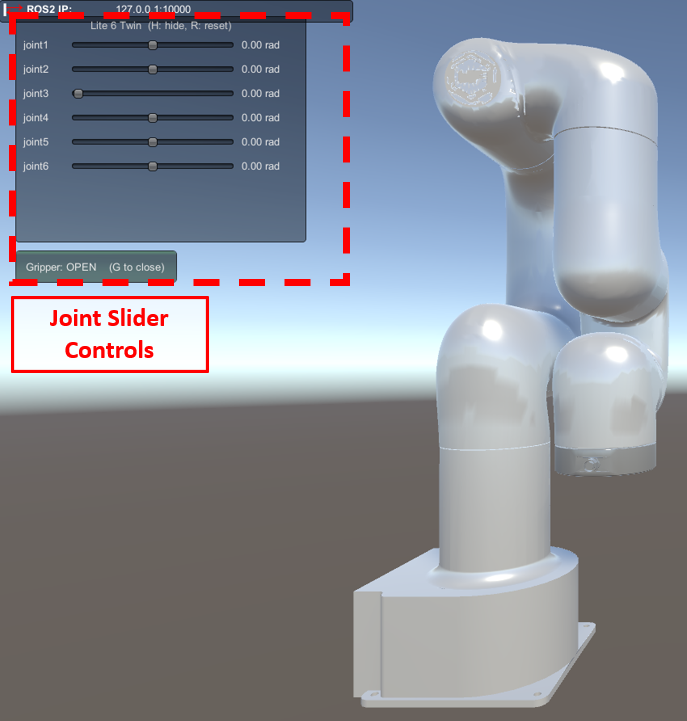}
    \caption{Digital twin in Unity}
    \label{fig:unity}
\end{subfigure}
\hfill
\begin{subfigure}[b]{0.48\columnwidth}
    \centering
    \includegraphics[width=\linewidth]{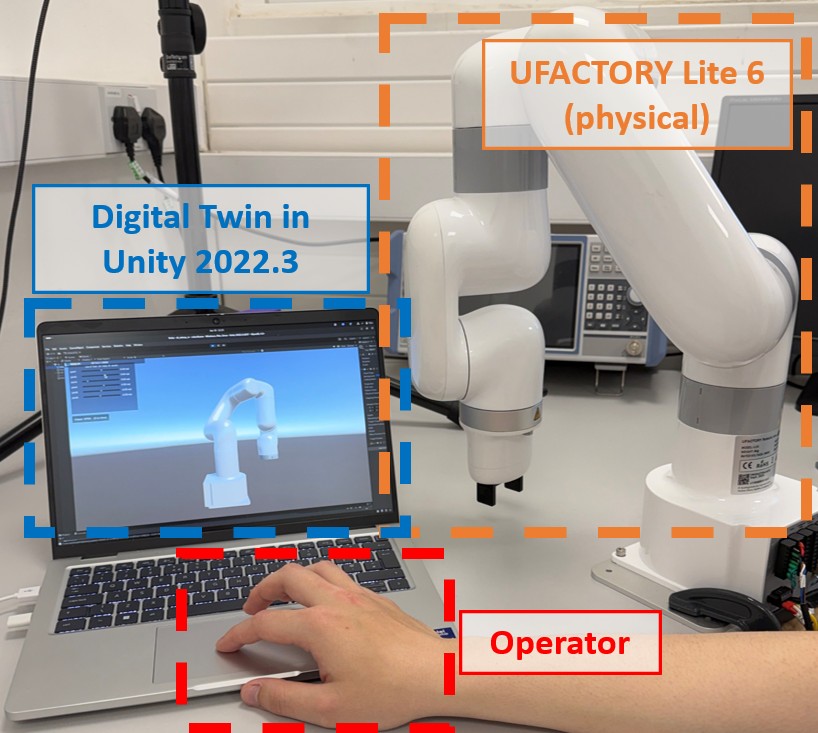}
    \caption{Physical robot}
    \label{fig:photo}
\end{subfigure}
\caption{DT testbed setup at the University of Warwick}
\label{fig:testbed2}
\end{figure}

The physical robot is a 6-DOF UFactory Lite~6 robot arm (Fig.~\ref{fig:photo}). A Linux host (Ubuntu 24.04, ROS~2 Jazzy) runs the xArm driver, a Unity 2022.3 environment for the DT, and a bridge that forwards the DT joint targets to the robot using the \texttt{set\_servo\_angle} service at \SI{5}{Hz} (Fig.~\ref{fig:unity}). Unity was chosen for its VR ecosystem and supported ROS~2 integration via ROS-TCP-Connector. Fig.~\ref{fig:testbed} summarizes the data flow. Network impairment is applied at the kernel level using \texttt{tc netem} on the ethernet interface connecting the host and the robot. Seven network profiles (Table~\ref{tab:profiles}) span typical deployment regimes (LAN, Wi-Fi, 4G, satellite). The AoI logger subscribes to the DT and real joint-state topics and records at \SI{20}{Hz} the active network profile, a wall-clock timestamp, and the elapsed time since the last received message on each channel. The receipt time is used rather than the source timestamp, due to the standard AoI definition. This ensures that the metric represents the impairment actually experienced by the DT. Post-processing scripts calculate the end-effector Cartesian position using forward kinematics, derive the instantaneous velocity by finite difference and evaluate the three metrics of Section~\ref{sec:metrics}.

\begin{table}[t]
\centering
\caption{Network profiles applied using \texttt{tc netem}.}
\label{tab:profiles}
\scriptsize
\begin{tabular}{lrrrr}
\toprule
Profile          & Delay (\si{ms}) & Jitter (\si{ms}) &
                   Loss (\%) & Rate (\si{Mbps}) \\
\midrule
Ideal            & 0   & 0   & 0   & --   \\
LAN              & 1   & 0.5 & 0   & 1000 \\
Good Wifi        & 10  & 5   & 0.1 & 100  \\
Congested Wifi   & 40  & 20  & 1.0 & 25   \\
4G               & 60  & 15  & 0.5 & 50   \\
Poor 4G          & 150 & 80  & 3.0 & 10   \\
Satellite        & 500 & 50  & 1.0 & 5    \\
\bottomrule
\end{tabular}

\end{table}

\subsection{Experimental process}
Each experimental trial is divided into eight sequential segments, with \SI{30}{s} allocated to each segment. All seven network profiles are traversed within a single trial, with the ideal condition repeated at both the beginning and end of the sequence to detect any drift in operator behavior or sensor bias over time, totaling \SI{4}{min} per trial. The operator moves sliders that move the DT joints, performing a disciplined metronome trajectory, with joint~1 oscillating over a \SI{60}{\degree} range at \SI{4}{s} period, joint~2 at \SI{6}{s} with \SI{40}{\degree} range, maintaining a consistent pace across all the network conditions. $N=11$ trials were collected across the entire experiment. A \SI{2}{s} period is removed after each profile change to allow for \texttt{netem} settling. The statistics for each profile are aggregated across all trials, with the two ideal segments pooled.
\vspace{-0.6\baselineskip}
\section{Results and Analysis}
\label{sec:results}

\subsection{Network condition statistics}
\label{sec:per-profile}

Table~\ref{tab:headline} reports the statistical performance across each network profile. As conditions degrade from the ideal to the satellite scenario, the 99th-percentile (P99) standard AoI increases from approximately \SI{200}{ms} (ideal) to \SI{1450}{ms}, which represents a \SI{7}{\times} degradation. Empirical ULAoI captures the tail more sharply: the probability of a bound violation ($P_\tau$) rises from $0.7\%$ to $80.6\%$ (a \SI{115}{\times} increase), while the conditional mean excess ($E_\tau$) rises from \SI{1}{ms} to \SI{454}{ms}. P95 K-AoI scales from approximately \SI{30}{mm} (ideal) to \SI{390}{mm} (satellite), a \SI{13}{\times} degradation. This degradation outpaces standard AoI growth because K-AoI compounds the temporal delay with the instantaneous velocity of the operator during latency spikes.
We show AoI at P99 and K-AoI at P95 to summarize each of the metrics near its information tail. AoI's distribution is shaped primarily by the network impairment, and  the standard high-percentile choice in networking and control is~P99. K-AoI's uppertail is more sensitive to single-sample motions, where brief high-velocity moments during network impairment spikes can produce isolated K-AoI values which are much larger than the operator's typical behavior. P95 captures most of the motion-weighted degradation while excluding the extreme single-samples.

\begin{table}[t]
\centering
\caption{Per-profile summary statistics and AoI/K-AoI rank correlation.
$N=11$ trials, $\tau=\SI{200}{ms}$.}
\label{tab:headline}
\scriptsize
\setlength{\tabcolsep}{4pt}
\begin{tabular}{lrrrrr}
\toprule
Profile         & AoI P99   & $P_\tau$ & $E_\tau$  & K-AoI P95 & Spearman $\rho$ \\
                & (\si{ms}) & (\%)     & (\si{ms}) & (\si{mm}) &        \\
\midrule
Ideal           & 199  & 0.7  & 1   & 31  & 0.30 \\
LAN             & 200  & 1.3  & 1   & 50  & 0.65 \\
Good Wifi       & 210  & 5.5  & 6   & 51  & 0.63 \\
Congested Wifi  & 296  & 22.3 & 36  & 71  & 0.63 \\
4G              & 280  & 24.8 & 36  & 64  & 0.66 \\
Poor 4G         & 639  & 53.2 & 129 & 129 & 0.34 \\
Satellite       & 1442 & 80.6 & 454 & 393 & 0.22 \\
\bottomrule
\end{tabular}
\end{table}
\subsection{Empirical Characterization of ULAoI and GPD Rejection}
\label{sec:gpd-fit}
The AoI tail above $\tau$ was tested to find whether it follows a Generalized Pareto Distribution (GPD) as Liao \emph{et~al.}~\cite{liao2023ulaoi} proposes. For each network profile, a GPD with maximum likelihood is fitted (with samples pooled across trials) and then a Kolmogorov-Smirnov (KS) test is applied against the fitted distribution. A KS-type goodness-of-fit diagnostic provides strong evidence that the fitted GPD does not adequately describe the AoI exceedance for most conditions.
The only exception is the \texttt{LAN} condition ($p = 0.78$), which nevertheless exhibits very few exceedances ($P_\tau = 1.3\%$), limiting the statistical power of the test rather than providing strong evidence of a valid fit. Across all other profiles, the estimated shape parameter remains mostly negative ($-0.87 \le \xi \le 0.05$), indicating bounded or light-tailed behavior rather than heavy-tailed regime assumed by the GPD model.
Two system-specific effects explain this mismatch: the bridge's deterministic \SI{5}{Hz} forwarding period dominates AoI even without impairment, producing a near-uniform, rather than heavy-tailed distribution; and \texttt{tc netem} introduces bounded delay and jitter distributions, which do not replicate the heavy-tailed latency characteristics commonly observed in real cellular or wide-area networks. Based on these results, we restrict the use of ULAoI in this study to its non-parametric components, namely the empirical exceedance probability $P_\tau$ and conditional excess mean $E_\tau$, both of which remain valid tail descriptors without requiring any distributional assumptions.

\subsection{K-AoI captures degradation invisible to AoI}
\label{sec:scatter}

A Spearman rank correlation is calculated for each sample between AoI and $\mathrm{K\text{-}AoI}_\mathrm{eef}$ within each network profile, across all $N=11$ trials pooled. If K-AoI were just a rescaling of AoI by a constant (or near constant) velocity, we would observe $\rho \to 1$. Substantially lower values would suggest that K-AoI carries independent information. Table~\ref{tab:headline} reports the correlations. All profiles exhibit $\rho \le 0.65$, with values dropping to $\rho = 0.22$ under satellite. None of the seven conditions approaches the correlation expected of a rescaled metric, indicating that K-AoI carries information not captured by AoI in any of the network conditions tested.

\begin{figure}[t]
\centering
\includegraphics[width=\columnwidth]{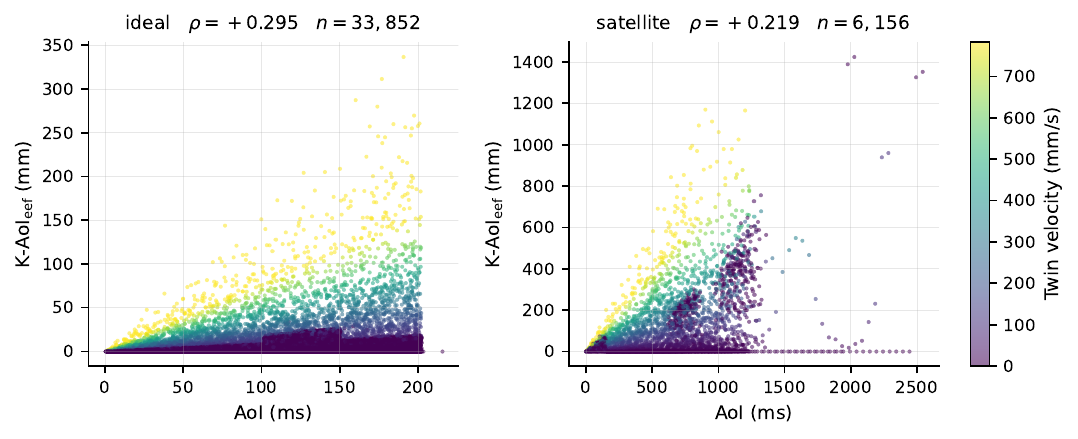}
\caption{Per-sample AoI versus K-AoI under ideal and
satellite conditions, pooled across $N=11$ trials. Points
colored by twin end-effector velocity. The horizontal band at K-AoI = 0 corresponds to stationary periods where AoI accumulates without spatial divergence, the upper envelope shows that moderate AoI at high operator velocity produces large K-AoI. Spearman $\rho = 0.295$ (ideal), $0.219$ (satellite).}
\label{fig:scatter}
\end{figure}

The divergence between AoI and K-AoI is most pronounced at the extremes of network conditions. Under ideal conditions ($\rho = 0.29$), AoI remains low and relatively uniform, while variations in velocity dominate spatial divergence; in this regime, K-AoI is weakly associated with AoI. Conversely, under satellite conditions ($\rho = 0.22$), AoI reaches larger values in the tail, with large delays occurring during both motion and stationary periods, weakening the relationship between the two metrics. Under moderate network impairments (LAN through 4G), the metrics partially align ($\rho \approx 0.6$), but the correlation remains insufficient to consider them redundant. We emphasize that K-AoI is not intended to replace AoI, but rather, each metric captures complementary aspects of synchronization quality, temporal freshness in the case of AoI, and motion-weighted spatial divergence in the case of K-AoI.
\vspace{-1.4\baselineskip}
\section{Future Work and Conclusion}
\label{sec:discussion}
We evaluate K-AoI, a velocity-weighted synchronization metric for digital twin teleoperation. Across $N=11$ trials on seven realistic network profiles, K-AoI correlates with AoI at $\rho \le 0.65$ throughout, dropping to $\rho = 0.22$ under satellite conditions, and K-AoI~P95 scales from \SI{30}{mm} (ideal) to \SI{390}{mm} (satellite), capturing velocity-modulated degradation that AoI cannot resolve and expressing it in meaningful units.

A KS test rejects the GPD hypothesis at $p < 10^{-3}$ for six of the seven conditions on the AoI tail above $\tau=\SI{200}{ms}$, with fitted shape parameters $\xi \in [-0.87, +0.05]$ indicating bounded rather than heavy tails, reflecting the bridge's deterministic \SI{5}{Hz} forwarding period. Whether ULAoI's GPD form remains useful at thresholds set significantly above one forwarding cycle, or when applied to residual jitter after subtracting the deterministic component, are open questions. The non-parametric ULAoI statistics ($P_\tau$ and $E_\tau$) characterize our seven conditions without such assumptions.

Two testbed extensions are planned. Replacing the slider interface with 6-DOF VR hand tracking would introduce better operator motion and let us evaluate K-AoI under more diverse velocity profiles. Integrating object tracking would extend spatial-divergence measurement from the end-effector to manipulated objects, broadening K-AoI's applicability to task performance assessment.
\vspace{-0.8\baselineskip}

\bibliographystyle{IEEEtran}
\bibliography{references}
\end{document}